\documentclass[10pt,twocolumn,letterpaper]{article}
\pdfoutput=1

\usepackage{iccv}
\usepackage{times}
\usepackage{epsfig}
\usepackage{graphicx}
\usepackage{amsmath}
\usepackage{amssymb}
\usepackage{framed}

\usepackage{multirow}
\usepackage[table,xcdraw]{xcolor} 
\usepackage{booktabs}
\usepackage{tikz,graphics,color,fullpage,float,epsf,caption,subcaption}
\newcommand{\tabincell}[2]{\begin{tabular}{@{}#1@{}}#2\end{tabular}} 
\usepackage{siunitx}
\usepackage[linewidth=1pt]{mdframed}
\usepackage[inkscapelatex=false]{svg}
\usepackage{microtype}      
\usepackage{epigraph}
\usepackage{soul}
\usepackage{fontawesome5}
\usepackage[numbers,sort]{natbib}
\usepackage{tabularx}
\usepackage[accsupp]{axessibility}

\usepackage{tcolorbox}

\usepackage{amssymb,mathrsfs,amsmath}

\newcommand{\AerialVLN}{AerialVLN \includegraphics[height=5mm]{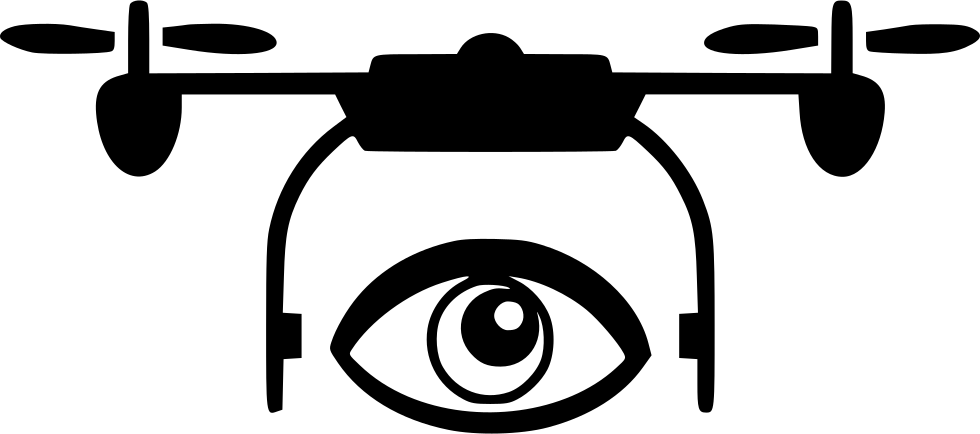}}

\usepackage{tikzpagenodes}

\usepackage {hyperref} 
\hypersetup {breaklinks=true,bookmarks=false}

\iccvfinalcopy 

\def\iccvPaperID{1991} 

\ificcvfinal\fi

\begin{document}

\title{\AerialVLN: Vision-and-Language Navigation for UAVs}

\author{
Shubo Liu$^{1\dagger}$ \and Hongsheng Zhang$^{1\dagger}$ \and Yuankai Qi$^{2}$ \and Peng Wang$^{1}$\thanks{Corresponding Author}
\and Yanning Zhang$^{1}$ \and Qi Wu$^{2}$ \\
$^1$Northwestern Polytechnical University\\
$^2$University of Adelaide\\
{\tt\small \{shubo.liu, hongsheng.zhang\}@mail.nwpu.edu.cn, qykshr@gmail.com,} \\
    {\tt\small \{peng.wang, ynzhang\}@nwpu.edu.cn, qi.wu01@adelaide.edu.au} 
}

\twocolumn[{%
\maketitle
\begin{figure}[H]
\hsize=\textwidth 
\centering
\vspace{-12mm}
\includegraphics[width=\textwidth]{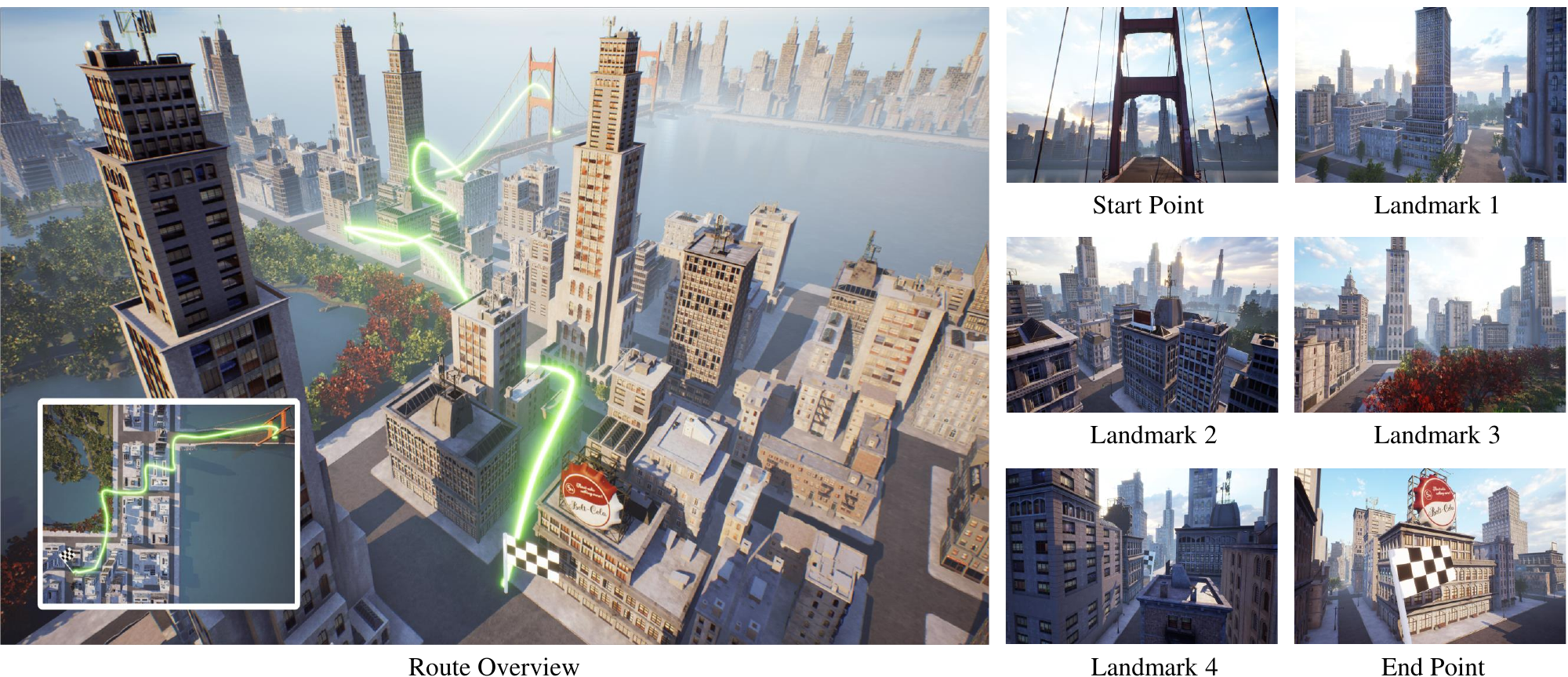}
\vspace{-7mm}
\begin{framed}
\vspace{-2mm}
    \caption*{\textit{\textbf{Instruction: }Take off, fly through the tower of cable bridge and down to the end of the road. Turn left, fly over the five-floor building with a yellow shop sign and down to the intersection on the left. Head to the park and turn right, fly along the edge of the park. March forward, at the intersection turn right, and finally land in front of the building with a red billboard on its rooftop.}}
\vspace{-6mm}
\end{framed}
\vspace{-5mm}
\captionof{figure}{An intelligent agent should be able to follow given natural language instructions and navigate to the destination in an unseen environment with visual perceptions along the way. The green line shows the agent's ground truth trajectory, and the chequered flag represents the end of it.}
\label{fig: Illustration of our dataset enviroment}
\end{figure}
\ificcvfinal\thispagestyle{empty}\fi
}]

\def\thefootnote{$\dagger$}\footnotetext{These authors contribute equally to this work}
\def\thefootnote{$\ast$}\footnotetext{Corresponding Author}

\begin{abstract}

Recently emerged Vision-and-Language Navigation (VLN) tasks 
have drawn significant attention in both computer vision and natural language processing communities.
Existing VLN tasks 
are built for agents that navigate on the ground, either indoors or outdoors.
However, many tasks require intelligent agents to carry out in the sky, such as UAV-based goods delivery, traffic/security patrol, and scenery tour, to name a few.
Navigating in the sky is more complicated than on the ground because agents need to consider the flying height and more complex spatial relationship reasoning.
To fill this gap and facilitate research in this field, we propose a new task named AerialVLN, which is UAV-based and towards outdoor environments.
We develop a 3D simulator rendered by near-realistic pictures of 25 city-level scenarios. 
Our simulator supports continuous navigation, environment extension and configuration. 
We also proposed an extended baseline model based on the widely-used cross-modal-alignment (CMA) navigation methods. We find that there is still a significant gap between the baseline model and human performance, which suggests AerialVLN is a new challenging task. 
Dataset and code is available at \url{https://github.com/AirVLN/AirVLN}.

\end{abstract}


\section{Introduction}
\label{sec:intro}

Recently, a bunch of vision-and-language navigation tasks, such as R2R~\cite{DBLP:conf/cvpr/AndersonWTB0S0G18}, RxR~\cite{DBLP:conf/emnlp/KuAPIB20}, REVERIE~\cite{DBLP:conf/cvpr/QiW0WWSH20}, TouchDown~\cite{DBLP:conf/cvpr/ChenSMSA19}, Alfred~\cite{DBLP:conf/cvpr/ShridharTGBHMZF20}, iGibson~\cite{li2021igibson, shenigibson, xia2020interactive}, have drawn a large amount of attention from different research communities like computer vision, natural language processing and robotics. %
These tasks as well as their datasets have greatly boosted the research of assembling the capabilities of vision and language understanding, cross-modality matching, path planning and reasoning~\cite{DBLP:conf/cvpr/HaoLLCG20,DBLP:conf/cvpr/KeLBHGLGCS19,DBLP:conf/eccv/QiPZHW20,DBLP:conf/iccv/ChattopadhyayHM21, DBLP:conf/cvpr/ChenGTSL22}. However, all these VLN tasks are designed for ground-based agents, which means agents can only navigate indoors or outdoors on the ground.
This overlooks another important application scenario: activities in the sky, which are becoming increasingly popular with the development of unmanned aerial vehicles (UAVs), especially multirotor.
We can now use UAVs to enjoy spectacular scenes without going out of houses and
they can be potentially utilized for goods delivery, traffic surveillance, search/rescue and security patrol~\cite{DJI_oil_and_gas,DJI_public_safety,DJI_aerial_surveying}.
To release humans from manually operating UAVs and to fill the research gap in the field of navigation in the sky, we propose {a city-level} UAV-based vision-and-language navigation task, named AerialVLN,  and a corresponding dataset. Navigating in the sky is significantly different from that on the ground in several aspects. 
\textbf{First}, AerialVLN has a larger action space. Compared to conventional ground VLN~\cite{DBLP:conf/cvpr/AndersonWTB0S0G18,DBLP:conf/emnlp/KuAPIB20,DBLP:conf/cvpr/ChenSMSA19,DBLP:conf/eccv/KrantzWMBL20,DBLP:conf/cvpr/QiW0WWSH20}, AerialVLN requires intelligent agents to additionally take actions such as ``rise up'' and ``pan down'' into consideration. Moreover, multirotors can move left/right without turning its head.
\textbf{Second}, the outdoor environments of AerialVLN are much bigger and more complex.  AerialVLN covers a large variety of city-level scenes. An intelligent agent is required to distinguish referred buildings/objects by their spatial relationship
from a bird-view as shown in Figure \ref{fig: Illustration of our dataset enviroment}. 
Although the TouchDown~\cite{DBLP:conf/cvpr/ChenSMSA19} task is also devised for outdoor navigation, its environments are static, while ours are interactive and dynamic. For example, our agent can land on a building, and the weather and illumination conditions can dynamically change in the environment.
\textbf{Third}, to mimic multirotor flying in real life, our AerialVLN has a much longer path than ground VLNs. On average, our AerialVLN involves a path length of 661.8 units\footnote{One unit equals one meter in our simulated city environment.}. %
There are about 9.7 referred objects in one instruction on average, which is more than 2.6 times as many as in the R2R dataset~\cite{DBLP:conf/cvpr/AndersonWTB0S0G18}.
\textbf{Fourth}, intelligent agents must learn to avoid getting stuck on objects {in 3D space}. This is more challenging than avoiding obstacles when navigating on the ground as in VLN-CE~\cite{DBLP:conf/eccv/KrantzWMBL20} because agents have to estimate the 3D shapes of obstacles and the distance to obstacles. All these new characters render AerialVLN a different and highly challenging task.

AerialVLN is implemented using Unreal Engine 4 \cite{UE4} and Microsoft AirSim plugins~\cite{DBLP:conf/fsr/ShahDLK17}, which enables continuous navigation and near-realistic rendering.
In total, we have collected 25 different city-level environments, covering a variety of scenes such as downtown cities, factories, parks, and villages, including more than 870 different kinds of objects. 
Our AerialVLN dataset consists of 8,446 flying paths obtained by experienced human UAV pilots who hold the AOPA (Aircraft Owners and Pilots Association) certificate. %
We pair each path with 3 instructions annotated by AMT workers in the standard dataset setting. Notably, we also align each sub-path to its sub-instruction, which enables fine-grained cross-modality matching learning.
On average, up to 83 words are in each instruction, involving a large vocabulary of 4,470 words.
Finally, we evaluate five baselines, including two golden standard VLN models in VLN, Seq2Seq model and cross-modal matching (CMA) model, and  our proposed model to serve as starting baselines on AerialVLN.

\begin{table*}[t]
\vspace{-2mm}
\centering
\resizebox{\linewidth}{!}{
\begin{tabular}{lccccccccc}
\hline
\textbf{Task} & \textbf{Routes} & \textbf{Instructions} & \textbf{Features}   & \textbf{Language} & \textbf{Action Space} & \textbf{Path Len.} & \textbf{Actions} & \textbf{Vocab} & \textbf{Intr. Len.} \\ \hline
R2R~\cite{DBLP:conf/cvpr/AndersonWTB0S0G18}           & 7,189           & 21,567                & Indoor, discrete    & Instruction       & Graph-based           & 10.0               & 5                & 3.1k           & 29                  \\
RxR~\cite{rxr}           & 13,992           & 13,992                  & Indoor, discrete    & Instruction       & Graph-based           & 14.9               & 8                & 7.0k           & 129                 \\
CVDN~\cite{DBLP:conf/corl/ThomasonMCZ19}          & 7,415           & 2,050                  & Indoor, discrete    & Dialog            & Graph-based           & 25.0               & 7                & 4.4k           & 34                \\
REVERIE~\cite{DBLP:conf/cvpr/QiW0WWSH20}       & 7k              & 21,702                & Indoor, discrete    & Instruction       & Graph-based           & 10.0                  & 5                & 1.6k           & 18                  \\
SOON~\cite{DBLP:conf/cvpr/ZhuL0YCL21}          & 40K             & 3,848                  & Indoor, discrete    & Instruction       & Graph-based           & 16.8                  & 9              & 1.6k           & 39                  \\
TouchDown\cite{DBLP:conf/cvpr/ChenSMSA19}     & 9,326           & 9,326                 & Outdoor, discrete   & Instruction       & Graph-based           & 313.9              & 35               & 5.0k           & 90                  \\
VLN-CE~\cite{DBLP:conf/eccv/KrantzWMBL20}        & 4,475           & 13,425                & Indoor, continuous  & Instruction       & 2  DoF                 & 11.1               & 56             & 4.3k           & 19                  \\
LANI~\cite{DBLP:conf/emnlp/MisraBBNSA18}          & 6,000           & 6,000                 & Outdoor, continuous & Instruction            & 2 DoF                     & 17.3                & 116                & 2.3k              & 57                \\
ANDH~\cite{Fan2022AerialVN}          & 6,269           & 6,269                 & Outdoor, continuous & Dialog            & 3 DoF                     & 144.7                & 7                & 3.3k              & 89                \\ \hline
AerialVLN        & 8,446           & 25,338                & Outdoor, continuous & Instruction       & 4 DoF                 & 661.8              & 204            & 4.5k           & 83                  \\
AerialVLN-S      & 3,916           & 11,748                & Outdoor, continuous & Instruction       & 4 DoF                 & 321.3              & 115            & 2.8k           & 82                  \\ \hline
\end{tabular}
}
\vspace{-3mm}
\caption{Comparison of existing vision-and-language navigation tasks. AerialVLN presents a city-level open environment dataset for aerial vision-and-language instruction-based navigation. Note that the en-US subset of RxR is considered for a fair comparison. Path length unit: meter.}
\vspace{-4mm}
\label{tab:DatasetComparisionStatistic}
\end{table*}

\section{Related Work}
\label{sec:relatedWork}
In this section, we review two types of closely related work: UAV  navigation and Ground-based VLN.

\noindent\textbf{{UAV Navigation.}}
Unmanned Aerial Vehicle (UAV) navigation has brought increasing attention in the last few decades.
Early UAV autonomous navigation requires solving the challenges of perceiving, mapping, localisation, decision-making (path-planning), action-decomposing and controlling. %
Inertial UAV navigation and GPS-based methods are commonly used together since the former might cause significant errors due to accumulation, and the latter is usually unable to localise the vehicle in high precision~\cite{Inertial_Navigation_Systems_for_UAV,DBLP:journals/sensors/LaconteKAVC22}. 
However, navigation in GPS-denied and unknown environments (such as cities with collapsed buildings or complex electromagnetic scenarios) becomes the bottleneck of intelligent UAVs. Vision-based navigation is then believed to be the solution to autonomous navigation~\cite{DBLP:journals/gsis/LuXXZ18,Vision_based_navigation_for_UAV,Courbon2010}.
The most similar works to ours are~\cite{DBLP:conf/rss/BlukisBBKA18,DBLP:conf/corl/BlukisMKA18,Fan2022AerialVN,DBLP:conf/emnlp/MisraBBNSA18} that language instructions are also provided. In~\cite{DBLP:conf/rss/BlukisBBKA18,DBLP:conf/corl/BlukisMKA18,DBLP:conf/emnlp/MisraBBNSA18},  a quadcopter agent is required to navigate by following natural language instructions in only one closed virtual field. The environment is of size 50$\times$50  with 6$\times$13 landmarks.
It has 6,000 instructions with an average length of 57 words and a vocabulary size 2,292.
Agents   are only allowed for horizontal movements (Forward, Left/Right and Stop).
By contrast, our AerialVLN is much larger, more complex, and closer to real-world scenarios: 
AerialVLN provides 25k crowd-sourced natural language instruction with an average length of 83 words and a vocabulary size 4,470. AerialVLN has 870 different kinds of objects and allows agents to move in 4-DOF as multirotor (Forward, Turn Left/Right, Ascent/Descend, Move Left/Right and Stop).
Moreover, AerialVLN presents 25 different open city-level environments, which enables intelligent agents to be trained and tested more comprehensively. 
Compared to ANDH~\cite{Fan2022AerialVN}, which focuses on dialogue-based aerial VLN with bird-view image input, our AerialVLN task requires agents to navigate with a first-person view and our environments are interactive and dynamic, which requires agents to learn to avoid obstacles. %
Regarding path length and data amount, our AerialVLN is four times of ANDH. More comparisons can be found in Table~\ref{tab:DatasetComparisionStatistic}.

\noindent\textbf{{Ground-based VLN Tasks.}}
A number of VLN tasks have been proposed for navigation on the ground.
Anderson~\etal~\cite{DBLP:conf/cvpr/AndersonWTB0S0G18} propose a Room-to-Room (R2R) navigation task, where given a detailed instruction, an agent is required to navigate from one room to another. 
Jain~\etal~\cite{DBLP:conf/acl/JainMKVIB19} propose to concatenate existing paths in R2R to form longer paths that are not the shortest ones between starting and ending points.
On the other hand, Chen ~\etal~\cite{DBLP:conf/cvpr/ChenSMSA19} propose an outdoor navigation task, TouchDown, to highlight challenges for outdoor environments.
Zhu~\etal~\cite{DBLP:conf/cvpr/ZhuL0YCL21} propose an object locating task, SOON, which uses detailed instruction descriptions.
By contrast, Qi~\etal~\cite{DBLP:conf/cvpr/QiW0WWSH20} propose a remote object grounding task, REVERIE, with concise, high-level instructions to better mimic the commands we humans give to each other.
However, all the above-mentioned VLN tasks are designed for intelligent agents navigating on the ground, as shown in Table \ref{tab:DatasetComparisionStatistic}.
This cannot reflect the challenges when navigating in the sky for intelligent agents like multirotor.
To address this problem, in this work, we propose a new task AerialVLN, which is designed for UAV navigation in the sky. 

\section{The AerialVLN Task}

As shown in Figure \ref{fig: Illustration of our dataset enviroment}, the proposed AerialVLN task requires an intelligent agent (a multirotor in virtual environments) to fly to the destination by following a given natural language instruction and its first-person view visual perceptions provided by the simulator. Unlike previous VLN tasks~\cite{DBLP:conf/cvpr/AndersonWTB0S0G18,DBLP:conf/emnlp/KuAPIB20,DBLP:conf/cvpr/ZhuL0YCL21}, we do not provide pre-build navigation graphs in our task, so any point not occupied by objects (such as buildings and trees) is navigable. This is closer to the practical scenario.

Formally, at the beginning of each episode, the agent is placed in an initial pose $P = [x, y, z, p, r, y']$, where $(x, y, z)$ denotes the agent's position and $(p, r, y')$ represents pitch, roll, yaw portion of the agent's orientation.
Then given a natural language instruction $X = <\omega_1, \omega_2, ... , \omega_L>$, where $L$ is the length of instruction and $\omega_i$ is a single word token, the agent is required to predict a series of actions.
The agent can take both the instruction and visual perceptions into consideration. 
Although our adopted simulator can provide panoramic observations, here we follow the most robotic navigation tasks ~\cite{DBLP:conf/eccv/KrantzWMBL20} setting to limit our baseline agent to the access of its front view perceptions (both depth and RGB images) $V_t = \{v_t^R, v_t^D\}$. The agent needs to rotate to obtain other views.
Navigation ends when the agent predicts a \textit{Stop} action or reaches a pre-defined maximum action number. The navigation is recognised as a success if the agent stops at a location that is less than 20 units to the target location, as 20 metres is the common size of a helipad in most countries. Considering the average size (radius) of our environment is around 3867.8 meters, this 20 metres landing area is rather small and challenging. 
The next section provides more details about the visual observation and action space.
\section{Simulator}

Our simulator is developed based on AirSim~\cite{DBLP:conf/fsr/ShahDLK17}  and Unreal Engine 4~\cite{UE4}. 
Below we detail its visual perceptions and  action space.

\noindent\textbf{Visual Observations.} In the  simulator, an embodied agent can move and observe in the continuous outdoor environment freely.
At each step $t$, the simulator outputs an RGB image $v_t^R$ and a depth image   $v_t^D$ of its front view.
Considering the outdoor environment setting, the depth sensor is allowed to perceive 100 meters ahead.
In addition, semantic segments are also accessible for future usage.
The simulator supports dynamic environments, such as blowing leaves, running cars, varying illumination (morning, noon, night) and different climate patterns (sun, rain, snow, fog). This can greatly narrow the gap when transferring trained agents to the real-world~\cite{DBLP:conf/corl/AndersonSTMPBL20}.

\noindent\textbf{Action Space.}
Although the simulator supports  flying towards any given direction and speed/distance, we consider the eight most common low-level actions in UAVs:
\textit{Move Forward}, \textit{Turn Left}, \textit{Turn Right}, \textit{Ascend}, \textit{Descend}, \textit{Move Left}, \textit{Move Right} and \textit{Stop}.
To balance the number of actions performed in a trajectory and the actual movement of the drone in an outdoor environment, the \textit{Move Forward} action   continuously moves 5 units along its current direction. The \textit{Move Left} and \textit{Move Right} actions continuously move 5 units along the corresponding direction, respectively. The \textit{Turn Left} and \textit{Turn Right} actions turn 15 degrees horizontally. The \textit{Ascend} and \textit{Descend} actions continuously move 2 units vertically.

\section{Dataset}
\label{sec:dataset}

In this section, we present the data collection policy and   analysis of collected instructions for AerialVLN.

\subsection{Data Collection}

\vspace{-1mm}

The data collection process contains two main steps: path generation and instruction collection. 
In contrast to R2R~\cite{DBLP:conf/cvpr/AndersonWTB0S0G18, DBLP:conf/corl/BlukisMKA18, DBLP:conf/corl/BlukisTNKA19}, of which the ground truth trajectories are automatically generated from navigation graph, 
we employed experienced multirotor manipulators (AOPA licensed) to complete the flying.
This enables the trained agent to learn real human remote pilots' behaviours.
To ensure the high quality of paths with a reasonable length, we ask human manipulators to pass several random-selected landmarks from a pre-defined landmark set including buildings, fountains, squares \etc.
In the simulator, we also provide hints about directions and distance to the next landmark to manipulators, which can help them better accomplish the task (refer to the supplementary for the interface). 
The output of the path generation step includes the multirotor's pose trace (a series of time-stamped 6-DoF multirotor poses).
The raw flying paths may have redundant motions,  
as manipulators sometimes need to look around to identify their positions and decide where to go. 
We remove such redundant motion for smoother ground truth trajectories.  
Then, the continuous paths are discretised into meta actions, such as ``turn left'' and ``move forward'' to enable training. %

For the second step, we use Amazon Mechanical Turk (AMT) to collect language instructions for these paths. Specifically, we show videos of drone flights and require the annotators to give natural language commands that can lead a pilot to complete the flying (refer to supplementary for the interface).
To enrich the language diversity and reduce bias, each video is annotated three times by different annotators. 
To avoid ambiguity caused by similar landmarks, referring expression is required,
such as ``land on the rooftop of the building near fountain''.
To validate data quality, all the collected instructions are manually checked by another group of workers. More details about the validation policy are presented in the supplementary material.

\begin{table*}[!t]
    \centering
    \vspace{-3mm}
    \resizebox{\linewidth}{1.9cm}{
        \begin{tabular}{l| c c| c c| c c| c c|c}\hline
                & \multicolumn{2}{c|}{R2R\cite{DBLP:conf/cvpr/AndersonWTB0S0G18}} & \multicolumn{2}{c|}{ANDH\cite{Fan2022AerialVN}} & \multicolumn{2}{c|}{TouchDown\cite{DBLP:conf/cvpr/ChenSMSA19}} & \multicolumn{2}{c|}{AerialVLN} & \\
            Phenomenon & $p$ & $\mu$ & $p$ & $\mu$ & $p$ & $\mu$ & $p$ & $\mu$ & Example in AerialVLN\\ \hline
            
            Reference & 100 &   3.7 &   92  &   1.9   &   100 &   9.2    &    100 &   \textbf{9.7}    & ...fly towards \textbf{the red bridge} across... \\
            
            Coreference &   32  &   0.5 &   8   &   0.1 &   60  &   1.1 &   68  &   \textbf{1.8} &\tabincell{c}{...move to the next building and after reaching \textbf{it} ...}\\
            
            Comparison &    4   &   0.0 &   32   &   0.4 &   12  &   0.1 &   20  &  0.2    &    ...get to the \textbf{tallest} tree in view...\\
            
            Sequencing &    16  &   0.2 &   8  &   0.1  &   84  &   1.6 &   68  &   \textbf{3.7} & ...go towards the \textbf{next} building and ...\\
            
            Allocentric Relation &  20  &   0.2 &   32  &   0.4 &   68  &   1.2 &   56  &   \textbf{4.6}    &...stop \textbf{on} the middle of the bridge... \\
            
            Egocentric Relation &   80  &   1.2 &   32  &   0.4 &   92  &   3.6 &   100 &   \textbf{7.1}    &...\textbf{stop when you get over the first tree}... \\
            
            Imperative &    100 &   4.0 &   100 &   1.1 &   100 &   5.2  &  100 &   \textbf{6.9}    &   \tabincell{c}{...\textbf{lift off} and \textbf{turn right} facing left of the old building and \textbf{head straight}...} \\
            
            Direction & 100 &   2.8 &   100  &   1.4    &   96  &   3.7 &   100 &   \textbf{4.6}    & ... \textbf{turn right and head back} into ... \\
            
            Temporal Condition &    28  &   0.4 &   20  &   0.2 &   84  &   1.9 &   76  &   \textbf{5.6}    &   ...look up \textbf{until} you see the sky... \\
            
            State Verification &    8   &   0.1 &   20  &   0.2 &   72  &   1.5 &   28  &   1.3 &   ...the road will now be \textbf{on your right}... \\\hline
        \end{tabular}
    }
    \vspace{-2mm}
    \caption{Linguistic phenomena in randomly sampled 25 instructions. AerialVLN task has a significant rate of reference, sequencing, spatial relationship and direction, which brings much more challenges to intelligent agents. $p$ and $\mu$ represent the percentage of instructions that present the phenomena and the average number of the phenomena appears in each instruction. For a fair comparison, TouchDown Navigation subset is used.}
    \label{tab:linguistic_phenomena}
    \vspace{-6mm}
\end{table*}

\subsection{Data Analysis}
\label{subsec:datasetanalysis}

\begin{figure}[!t]
    \centering
    \subfloat[\label{fig:a} Word cloud of nouns]{
		\includegraphics[width=0.49\linewidth]{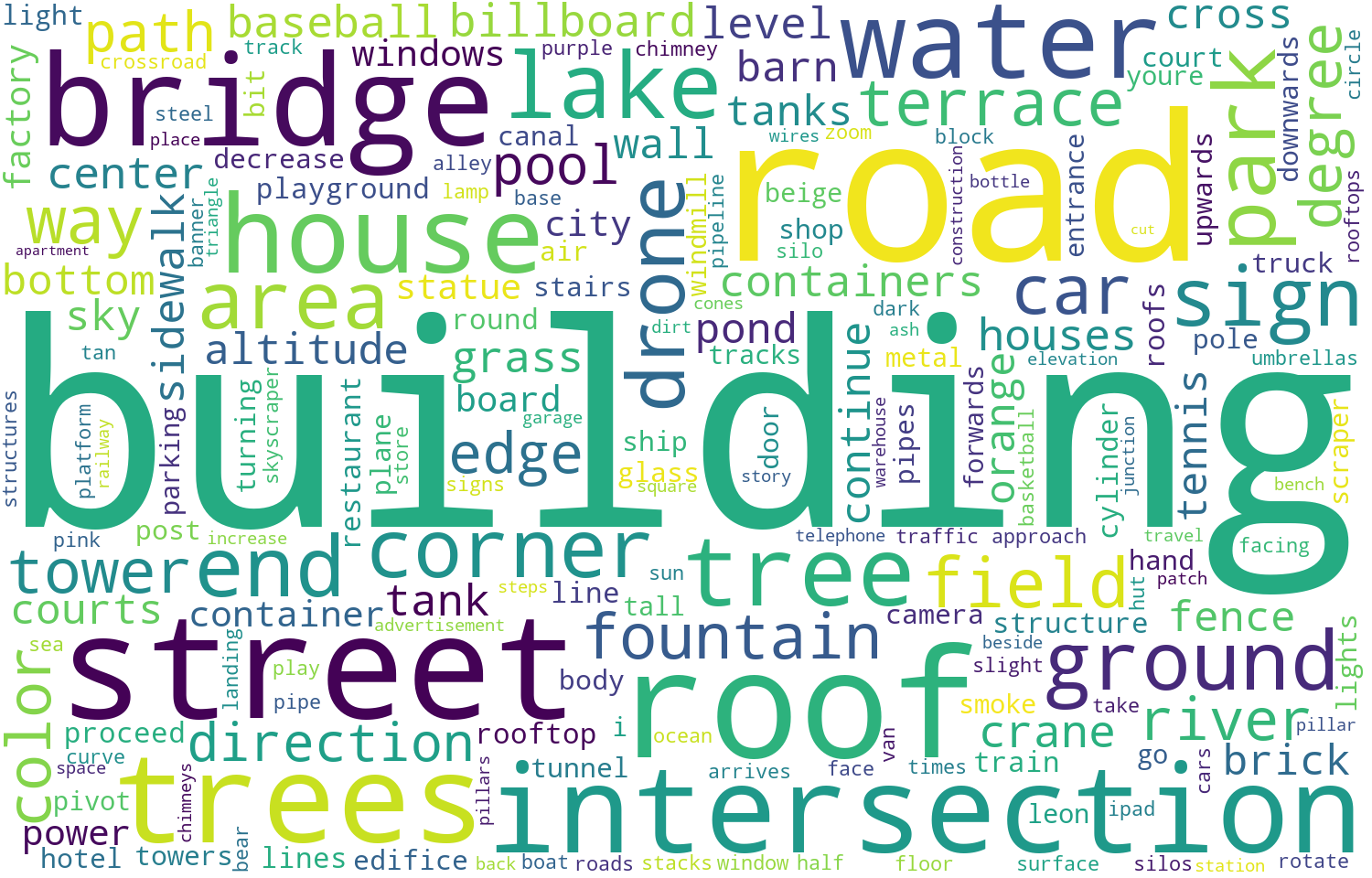}}
    \subfloat[\label{fig:b} Word cloud of verbs]{
		\includegraphics[width=0.49\linewidth]{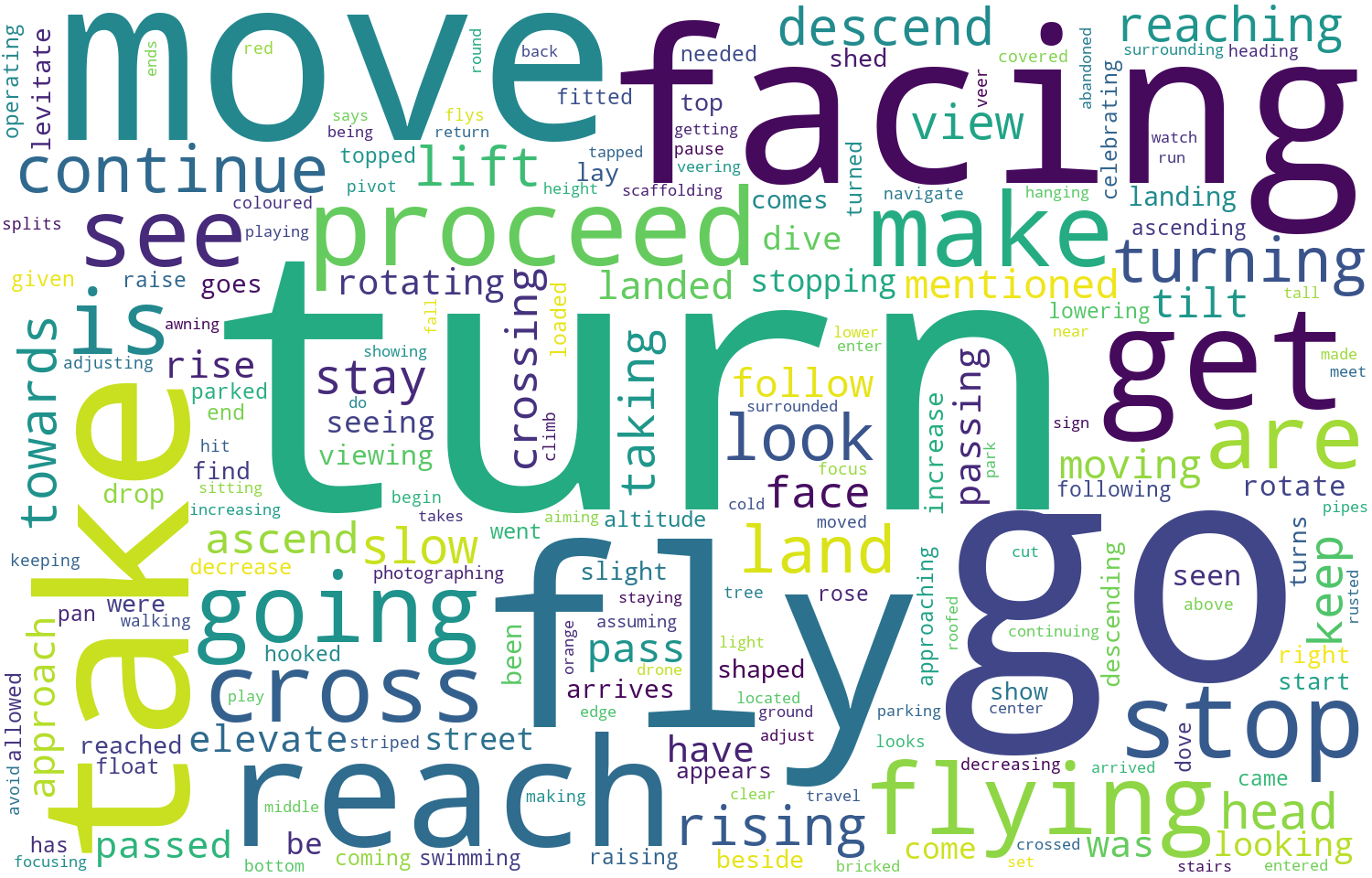}}
    \vspace{-3mm}
    \caption{Statistics of nouns and  verbs.}
    \vspace{-6mm}
    \label{fig:word_cloud}
\end{figure}

We totally collected 25,338 instructions with a vocabulary of 4,470 words. 
On average, each instruction has 83 words. 
Figure \ref{fig:word_cloud} presents the relative word frequency in the form of word cloud, where the larger the font, the more frequently the word is used.
Figure \ref{fig:word_cloud}(a) shows 
that   ``building'' and ``road'' are mostly used as reference objects for navigation. 
Figure \ref{fig:word_cloud}(b) shows 
that ``turn'', ``go'' and ``fly'' are the most common verbs.

\begin{figure}[!htbp]
    \centering
    \subfloat[\label{fig:a} Instruction length]{
		\includegraphics[width=0.48\linewidth]{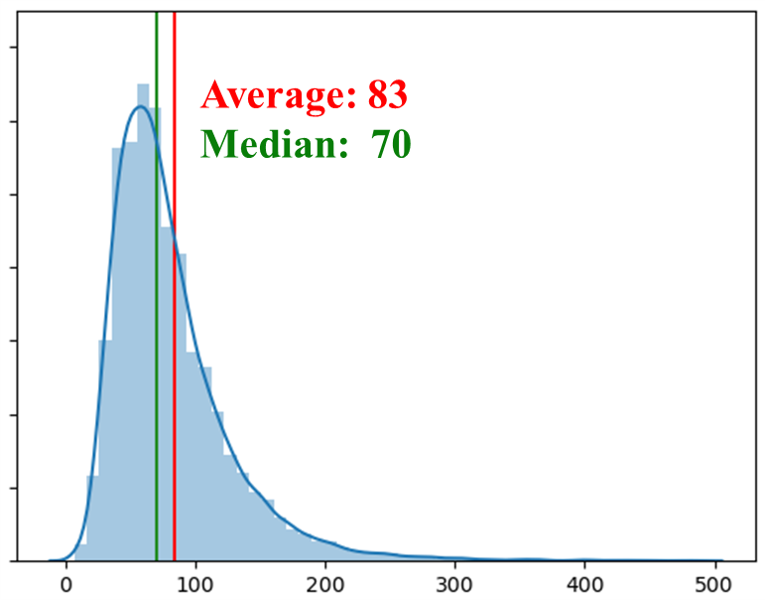}}%
  \hfill
    \subfloat[\label{fig:b} Number of actions per path]{
		\includegraphics[width=0.48\linewidth]{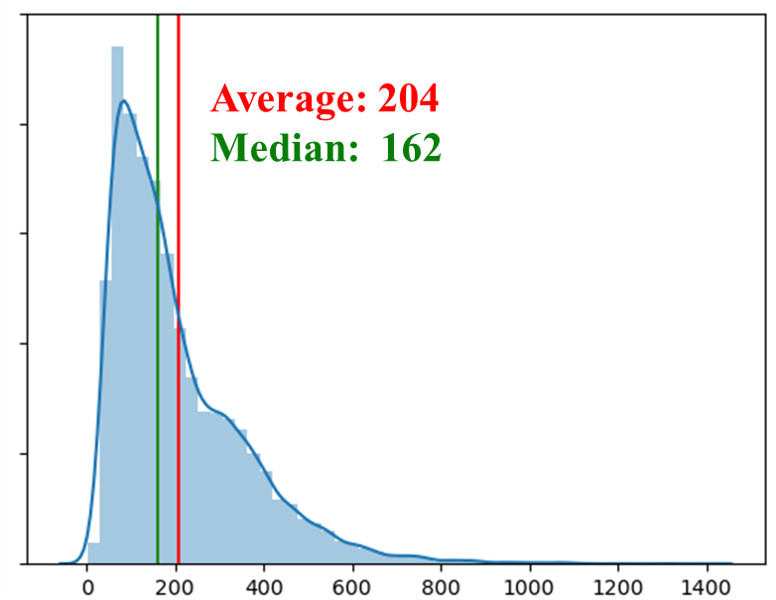}}%
    \vspace{-3mm}
    \caption{Instruction length and   number of actions.
    }
    \label{fig:instruction_action_ditribution}
\end{figure}

In Table \ref{tab:DatasetComparisionStatistic}, we provide a comparison between our AerialVLN dataset and other popular VLN datasets. 
It shows that AerialVLN has the largest average path length, which is about five times ANDH and 40 times the ground-based VLN, RxR.
At the same time, our dataset has the most average actions per path, which is about four times VLN-CE and six times TouchDown.
In terms of the number of instructions, our dataset is about twice VLN-CE and RxR, four times ANDH. 
All these characters render our dataset extremely challenging. 
In Figure~\ref{fig:instruction_action_ditribution}, we present the distribution of instruction length and the number of actions. As shown in Figure~\ref{fig:instruction_action_ditribution}(a),  instruction length ranges from 50 to 130 words. Figure ~\ref{fig:instruction_action_ditribution}(b) shows that most paths have 50 $\sim$ 240 actions. These diversities make AerialVLN more challenging. %

As in previous work~\cite{DBLP:conf/cvpr/ChenSMSA19,DBLP:conf/emnlp/KuAPIB20}, we have also conducted statistics on linguistic phenomena of randomly sampled 25 instructions with comparison to other VLN datasets in Table \ref{tab:linguistic_phenomena}. 
It shows that our AerialVLN task has a significant rate of reference, coreference, sequencing, spatial relationship and direction, which brings much more challenges. 

\noindent\textbf{Dataset Split.} Following the common practice in the VLN community, we divide our dataset into the train, val\_seen, val\_unseen, and test splits. The word ``seen'' means the visual environments that have been seen in the train split. 
As shown in Table~\ref{tab:dataset_split}, we assign $17$ scenes for training and val\_seen split, where the train set contains $16,380$ instructions from $5,460$ paths.
For val\_seen, we assign $1,818$ instructions from $606$ paths.
The val\_unseen and the test split are both assigned $8$ scenes, but the test split is about double the size of the val\_unseen split, including $1610$ paths with $4,830$ instructions. Please note that the test split is built on unseen scenes as well and the goal locations for the test set will not be released.
Instead, we provide an evaluation server where UAV trajectories can be uploaded for scoring.

Besides the standard dataset setting (with all 25 scenes), we also present a variant for small scenes, AerialVLN-S.
It preserves the same split, but it has 17 scenes with a smaller scale and evenly-distributed path length, which results in a shorter path length (average path length reduces 51.5\%) and shorter instruction length. 
In AerialVLN-S setting, the agent can sometimes even observe the goal at starting point.
In this variant, there are $10,113$ instructions for the train and $333$ for validation seen splits, respectively.
$531$ instructions are assigned for validation unseen and $771$ instructions for the test set, as shown in Table~\ref{tab:dataset_split}.
We hope future researchers employ AerialVLN to tackle long path length 3D VLN tasks in the unseen environment and focus on the investigation of action learning in long time horizons and sparse reward; while utilising AerialVLN-S as the benchmark for general 3D aerial VLN tasks in first-person view.

\begin{table}[htbp]
\centering
\resizebox{0.99\linewidth}{!}{
\begin{tabular}{l|ccc|ccc}
\hline
\multirow{2}{*}{} & \multicolumn{3}{c|}{AerialVLN} & \multicolumn{3}{c}{AerialVLN-S} \\
                  & Scene    & Path     & Instr.   & Scene     & Path    & Instr.    \\ \hline
Train             & 17       & 5,460    & 16,380   & 12        & 3,371    & 10,113     \\
Val Seen          & 17       & 606      & 1,818     & 12        & 111     & 333       \\
Val Unseen        & 8        & 770      & 2,310     & 5         & 177     & 531       \\
Test              & 8        & 1,610     & 4,830     & 5         & 257     & 771       \\ \hline
\end{tabular}
}
\caption{Dataset splits. AerialVLN is the full dataset, and AerialVLN-S is for small scenes.}
\label{tab:dataset_split}
\vspace{-3mm}
\end{table}

\section{Experiment and Results}
\label{sec:exp}

\begin{figure*}[!t]
    \vspace{-1mm}
    \centering
    \includegraphics[width=\linewidth]{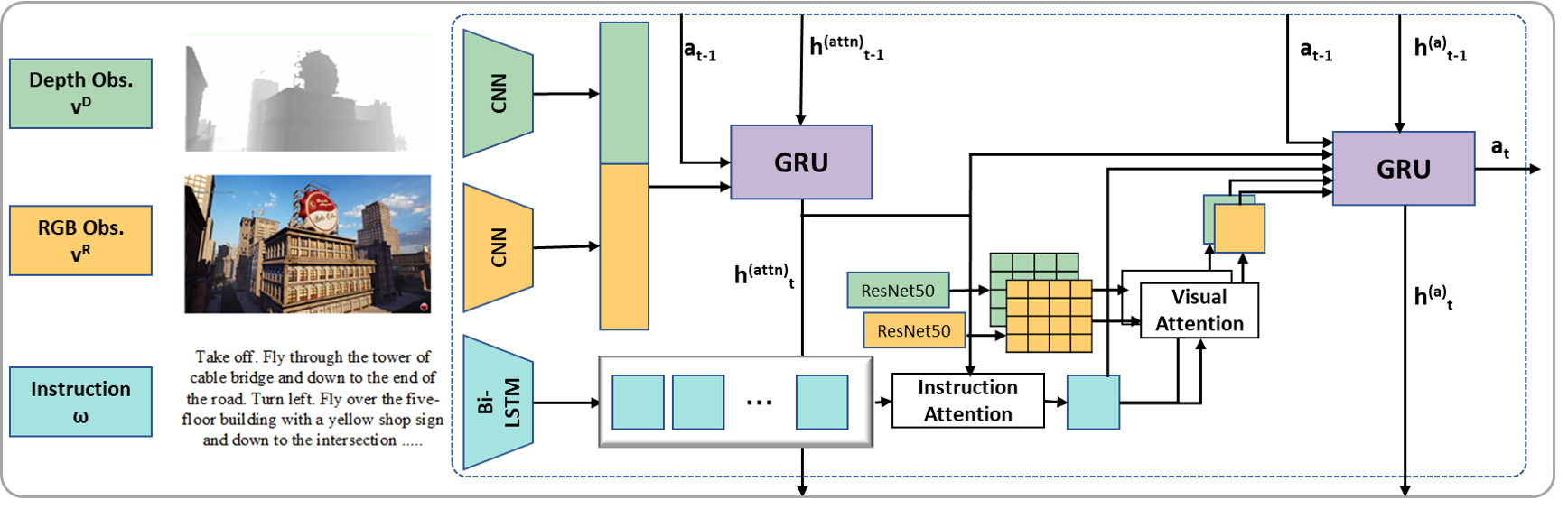}
    \vspace{-8mm}
    \caption{Main architecture of the Cross-Modal  Attention model}
    \vspace{-6mm}
    \label{fig:models-cma}
\end{figure*}

In this section, we first present the evaluation metrics and training details of baseline models. 
Then we provide extensive evaluation and analysis.

\subsection{Evaluation Metrics}
\label{sec: evaluation metrics}

We adopt four widely used metrics in VLN tasks~\cite{DBLP:conf/cvpr/AndersonWTB0S0G18,DBLP:conf/emnlp/KuAPIB20,DBLP:conf/cvpr/QiW0WWSH20,DBLP:conf/acl/JainMKVIB19}:
Success Rate (SR), where one navigation is considered successful if the agent stops within 20 meters of the destination;
Oracle Success Rate (OSR), where one navigation is considered oracle success if the distance between the destination and any point on the trajectory is less than 20 meters;
Navigation Error (NE), the distance between the stop location to the destination;
Success rate weighted by Normalised Dynamic Time Warping (SDTW), which considers both the navigation success rate and the similarity between ground truth path and model predicted path ~\cite{DBLP:conf/nips/IlharcoJKIB19}.

\subsection{Results}

We evaluate five baseline models on our task. Four of these baselines have served as a golden standard for VLN tasks as in~\cite{DBLP:conf/cvpr/AndersonWTB0S0G18,DBLP:conf/corl/AndersonSTMPBL20,DBLP:conf/cvpr/QiW0WWSH20}. The other baseline is  our extension to the best existing baseline. Below we first briefly introduce baselines and then present the results.

\subsubsection{Baselines}
\label{sec:Baselines}

\textbf{Random.}
The agent randomly selects actions at each location and stops until the `stop' action is selected or when reaching the max steps. This is widely used to reflect how big the solution space can be.

\noindent\textbf{Action Sampling.}
Action Sampling agents explore the statistical characteristic of the dataset by sampling actions according to the action distribution of the training set. 
This can be used to measure the similarity of the action distribution on   evaluation and training splits.

\noindent\textbf{LingUNet.}
LingUNet~\cite{DBLP:conf/emnlp/MisraBBNSA18} is a baseline model used by previously aerial VLN task LANI.
Consider that LANI assume that agent can see the destination from the start point, LingUNet has an episode-wise paradigm. 
However, such assumption can't stand in AerialVLN task, we thus adapt LingUNet model into a step-wise paradigm. 

\noindent\textbf{Sequence-to-Sequence.}
Seq2Seq~\cite{DBLP:conf/cvpr/AndersonWTB0S0G18} is a  baseline model with a recurrent policy. It takes as input the  \ concatenation of \ the \ RGB \ feature  $\overline{v_t}^R =\mathrm{meanpool}(\mathrm{ResNet}_{RGB}(v_t^R))$, \ the \ Depth \ feature $\overline{v_t}^D = \mathrm{ResNet}_{Depth}(v_t^D)$, 
and the instruction embedding $s = \mathrm{LSTM}(\omega_1, ..., \omega_L)$. Then it   projects them to a hidden representation $h_t^{(a)} = \mathrm{GRU}([\overline{v_t}^R, \overline{v_t}^D, s], h_{t-1}^{(a)})$, which is further used to predict a distribution over the action space. The one with the largest probability is selected as the next action  $a_t = \underset{a}{\mathrm{argmax}} \;\mathrm{softmax}(W_a h_t^{(a)} + b_a)$.

\begin{table*}[!tbp]
\centering
\begin{subtable}[Performance of AerialVLN task]{\linewidth}{
\resizebox{\linewidth}{!}{%
\begin{tabular}{p{0.3cm}<{\centering}p{2.5cm}|cccc|cccc|cccc}
\hline
                              & \multicolumn{1}{c|}{}                                     & \multicolumn{4}{c|}{\textbf{Validation Seen}}                                 & \multicolumn{4}{c|}{\textbf{Validation Unseen}}                               & \multicolumn{4}{c}{\textbf{Test Unseen}}                                     \\
\multirow{-2}{*}{\textbf{\#}} & \multicolumn{1}{c|}{\multirow{-2}{*}{\textbf{AerialVLN}}} & NE/m $\downarrow$ & SR/\% $\uparrow$ & OSR/\% $\uparrow$ & SDTW/\% $\uparrow$ & NE/m $\downarrow$ & SR/\% $\uparrow$ & OSR/\% $\uparrow$ & SDTW/\% $\uparrow$ & NE/m $\downarrow$ & SR/\% $\uparrow$ & OSR/\% $\uparrow$ & SDTW/\%$\uparrow$ \\ \hline
\rowcolor[HTML]{EFEFEF} 
1                             & Random                                                    & 300.8             & 0.0              & 0.0               & 0.0                & 351.0             & 0.0              & 0.0               & 0.0                & 356.3             & 0.0              & 0.0               & 0.0               \\
2                             & Action Sampling                                           & 383.1             & 0.1              & 2.1               & 0.1                & 434.9             & 0.2              & 2.1               & 0.1                & 441.9             & 0.2              & 1.8               & 0.1               \\
\rowcolor[HTML]{EFEFEF} 
3                             & Seq2Seq                                                   & 480.4             & 2.9              & 10.2              & 1.0                & 551.8             & 1.1              & 5.6               & 0.3                & 558.8             & 1.0              & 4.9               & 0.3               \\
4                             & CMA                                                       & 293.5             & 2.3              & 6.5               & 0.8                & 360.7             & 1.6              & 4.4               & 0.5                & 358.6             & 1.6              & 4.1               & 0.5               \\
\rowcolor[HTML]{EFEFEF} 
5                             & Human                                                     & -                 & -                & -                 & -                  & -                 & -                & -                 & -                  & 73.5              & 80.8            & 80.8              & 14.2                \\ \hline
\end{tabular}
}
}
\end{subtable}
\begin{subtable}[Performance of AerialVLN-S task]{\linewidth}{
\resizebox{\linewidth}{!}{%
\begin{tabular}{p{0.3cm}<{\centering}p{2.5cm}|cccc|cccc|cccc}
\hline
                              & \multicolumn{1}{c|}{}                                       & \multicolumn{4}{c|}{\textbf{Validation Seen}}                                 & \multicolumn{4}{c|}{\textbf{Validation Unseen}}                               & \multicolumn{4}{c}{\textbf{Test Unseen}}                                      \\
\multirow{-2}{*}{\textbf{\#}} & \multicolumn{1}{c|}{\multirow{-2}{*}{\textbf{AerialVLN-S}}} & NE/m $\downarrow$ & SR/\% $\uparrow$ & OSR/\% $\uparrow$ & SDTW/\% $\uparrow$ & NE/m $\downarrow$ & SR/\% $\uparrow$ & OSR/\% $\uparrow$ & SDTW/\% $\uparrow$ & NE/m $\downarrow$ & SR/\% $\uparrow$ & OSR/\% $\uparrow$ & SDTW/\% $\uparrow$ \\ \hline
\rowcolor[HTML]{EFEFEF} 
S1                            & Random                                                      & 109.6             & 0.0              & 0.0               & 0.0                & 149.7             & 0.0              & 0.0               & 0.0                & 148.5             & 0.0              & 0.0               & 0.0                \\
S2                            & Action Sampling                                             & 213.8             & 0.9              & 5.7               & 0.3                & 237.6             & 0.2              & 1.1               & 0.1                & 242.0             & 0.7              & 2.5               & 0.3                \\
\rowcolor[HTML]{EFEFEF} 
S3                            & LingUNet                                                 & 383.8             & 0.6              & 6.9              & 0.2                & 368.4             & 0.4              & 3.6              & 0.9                & 399.8             & 0.1              & 3.1               & 0.1                \\
S4                            & Seq2Seq                                                     & 146.0             & 4.8              & 19.8              & 1.6                & 218.9             & 2.3              & 11.7              & 0.7                & 214.6             & 2.2              & 9.4               & 0.7                \\
\rowcolor[HTML]{EFEFEF} 
S5                            & CMA                                                         & 121.0             & 3.0              & 23.2              & 0.6                & 172.1             & 3.2              & 16.0              & 1.1                & 178.5             & 3.9              & 13.1              & 1.4                \\
S6                            & Seq2Seq-DA                                                  & 85.5              & 9.9              & 24.1              & 4.5                & 143.5             & 4.0              & 10.9              & 0.7                & 140.2             & 3.5              & 9.5               & 0.6                \\
\rowcolor[HTML]{EFEFEF} 
S7                            & CMA-DA                                                      & 92.2              & 9.9              & 26.5              & 3.7                & 122.7             & 4.5              & 13.9              & 1.0                & 125.4             & 4.3              & 14.8              & 1.2                \\
S8                            & Ours (LAG)                                               & 90.2              & 7.2              & 15.7              & 2.4                & 127.9             & 5.1              & 10.5              & 1.4                & 128.3             & 4.5              & 11.6              & 1.3                \\ \hline
\end{tabular}
}
}
\end{subtable}
\vspace{-3mm}
\caption{Performance of baselines on our AerialVLN task (Row 1-5) and AerialVLN-S task (Row S1-S7).  There is a significant gap to human performance. 
}
\vspace{-3mm}
\label{tab:baseline_performance_compare}
\end{table*}

\noindent\textbf{Cross-Modal Attention.}
Cross-Modal Attention (CMA) is a classical baseline model for  VLN tasks.
As shown in Figure \ref{fig:models-cma}, the CMA baseline is based on a bi-directional LSTM and divides the whole process into two parts. 
One is tracking visual observations,  and the other one is decision-making. 
The former is formulated as $h_t^{(attn)} = \mathrm{GRU}([ \overline{v_t}^R, \overline{v_t}^D, a_{t-1}], h_{t-1}^{(attn)})$, where $a_{t-1} \in \mathbb{R}^{1\times32} $ and is a learned linear embedding of the previous action. 
The latter encodes the instruction embedding first and outputs the intermediate hidden state $S = \{s_1, ... , s_L\} = \mathrm{BiLSTM}(\omega_1, ... , \omega_L)$. 
Then the attention mechanism is applied to instructions and images, which is $\hat{s}_t = \mathrm{Attn}(S, h_t^{(attn)})$, $\hat{v}_t^R = \mathrm{Attn}(v_t^R, \hat{s}_t)$, $\hat{v}_t^D = \mathrm{Attn}(v_t^D, \hat{s}_t)$, where $\mathrm{Attn}$ is a scaled dot-product attention. 
Finally, all these features and embeddings are concatenated to serve as the input of the recurrent network and predict an action for the agent to execute. 

\begin{figure}[!htbp]
    \centering
    \subfloat[\label{fig:lookAheadGuidance_a} Shortest path guidance]{
		\includegraphics[width=0.49\linewidth]{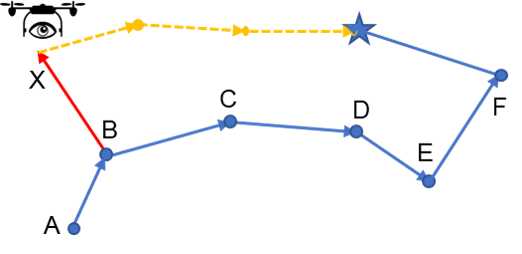}}%
    \subfloat[\label{fig:lookAheadGuidance_b} Look-ahead guidance]{
		\includegraphics[width=0.5\linewidth]{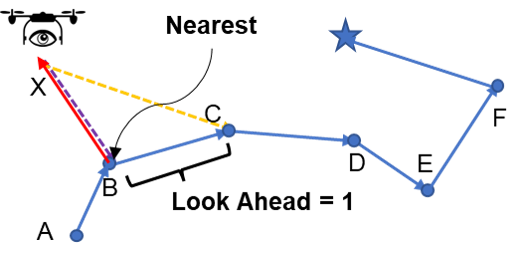}}
    \vspace{-1mm}
    \caption{Illustration of Look-ahead Guidance. `A' denotes starting location; `$\star$' denotes destination; `X' denotes  current location; Blue path denotes ground-truth;
    Yellow path denotes ``generated ground-truth'' when the agent deviates from the real ground-truth path.
    }
    \label{fig:lookAheadGuidance}
\end{figure}

\noindent\textbf{Look-ahead Guidance (LAG).}
When training models in a student-forcing fashion, the ground-truth actions are usually determined according to the shortest path from the current location to the destination (see Figure~\ref{fig:lookAheadGuidance}(a)) in most existing methods. However, this is unreasonable because the instructions do not describe the shortest path from starting to the destination.
To mitigate this issue, we inspired by \cite{DBLP:conf/emnlp/RaychaudhuriWPJ21} and propose a new strategy that generates ground-truth actions according to a ``look-ahead'' path. As shown in Figure \ref{fig:lookAheadGuidance}(b), assuming the agent is at location X currently, the look-ahead path is determined by three steps: (1) find the shortest path to return to the ground-truth path (X$\rightarrow$B in the example); (2) navigate along the ground-truth path 10 steps {(look-ahead step = 10)}, assuming arrive at location C; (3) the look-ahead path is the shortest path from X  to location C, and the ground-truth action for the next step is the first step on this path. We combine the aforementioned CMA model and our look-ahead guidance as our new baseline, denoted as LAG.

\subsubsection{Results}

Table ~\ref{tab:baseline_performance_compare} shows all the results.
The results show that:

\noindent 1. Random action hardly succeeds. The success rate of the Random model (Row 1 and S1) is 0\%. Even if we sample actions according to the action distribution of the training split, the success rate still remains below 1\% (Row 2 and S2). Moreover, the oracle success rate is always below 3\% on unseen splits. This indicates an agent can hardly reach even passing by the destination if it cannot understand the instructions, visual perceptions, and their alignment.

\noindent 2. LingUNet achieves limited success. Performance on Unseen cases only slightly better than Action Sampling (Row S2$\sim$S3). This may be attributed to the lack of recurrent structures in the decision component of the Baseline, resulting in the model's inability to effectively model historical information.

\noindent 3. The golden baselines Seq2Seq and CMA achieve success rate 1.0\%$\sim$ 1.6\% on unseen splits of the full dataset (Val\_Unseen and Test\_Unseen, Row 3$\sim$4) and 2.2\%$\sim$3.9\% on the AerialVLN-S dataset (Row S4$\sim$S5). At the same time, the oracle success rate also rises to 5\% and 16\%, respectively. This indicates that learning-based models have a larger chance to succeed than random models. However, the success rate is still rather low compared to human performance (SR: $\sim$80\%). 

\noindent 4. When applying the Dataset Aggregation (DA~\cite{DBLP:journals/jmlr/RossGB11}, an offline student-forcing strategy where executed actions are sampled from predictions instead of ground-truth actions) technique to mitigate the training-test disconnection problem (agents in test are not exposed to the consequences of their actions during training), the performance becomes better with about 6\% improvement on seen split and about 1\% improvement on unseen splits (Row S6$\sim$S7). This demonstrates exploring non-ground-truth actions helps learn more from training data and increases generalisation ability. On the other hand, comparing to the results of the same model (\eg, CMA-DA) on ground-based VLN tasks, such as continuous R2R, the performance on AerialVLN is rather low: SR 4.5\% \vs {27\%}. This indicates AerialVLN is much more challenging.

\noindent 5. By incorporating our proposed look-ahead guidance (LAG) to the best baseline model CMA, it achieves further enhanced performance (Row S8) on unseen splits in terms of both SR and SDTW, which demonstrates the look-ahead guidance can help the agent to fly according to instructions.

\begin{figure*}[htb]
    \centering
    \includegraphics[width=\linewidth]{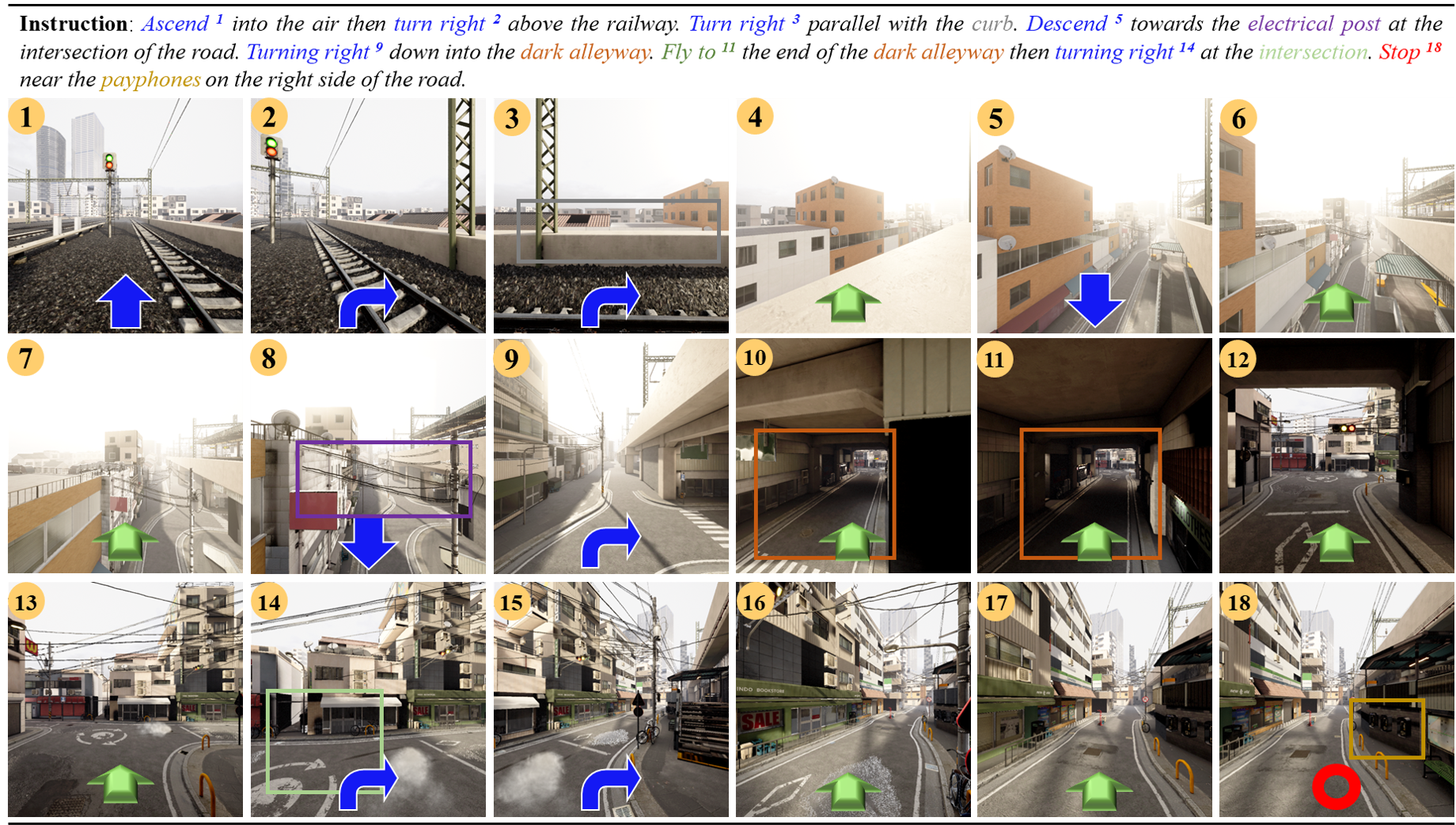}
    \vspace{-6mm}
    \caption{Visualisation of a successful navigation of our LAG model. 
    Green arrows indicate horizontal movement motions (Move Forward, Move Left/Right); blue arrows represent vertical motion (Move Up/Down) and horizontal rotation (Turn Left/Right). 
    The final red circle denotes Stop. 
    We highlight  aligned landmarks   by coloured bounding boxes in images and words in the instruction using the same colour.
    The superscript of words denotes the index of the corresponding action in images.
    }
    \vspace{-5mm}
    \label{fig:case_study_visualization}
\end{figure*}

We also present a qualitative result in Figure \ref{fig:case_study_visualization}. It shows that when the agent can align visual and textual landmarks (as well as understand rotation commands), it has a large chance to succeed.

\noindent\textbf{Possible reasons for failures} 
We find that the length of path magnificently influences the success rate. Seq2Seq and CMA could follow instructions at an early stage but they cannot get back on track once deviate. Take AerialVLN for example, we further divide it into a long-path set (average path length 813.2m) and a short-path set (average path length 326.9m). Success rate on the former is only 1.8\% while the latter can be up to 7.4\%.
Failure to stop correctly also leads to the low success rate. 
Supplementary material provides further failure analysis. %
As shown in Table \ref{tab:baseline_performance_compare}, OSR of both Seq2Seq and CMA is significantly higher than SR, which suggests that the agent has passed the goal location and failed to stop around it. 

\subsection{Modality Ablation Study}

To investigate the importance of different modalities in this task,  we conduct an ablation study based on the CMA model via removing RGB, Depth, RGB+Depth (Vision), and Language from inputs, one by one. 
The results are presented in Table \ref{tab:ablation}. 

\begin{table}[htbp]
\vspace{-2mm}
\centering
\resizebox{\linewidth}{!}{
\begin{tabular}{lcc|cccc}
\hline
\multirow{2}{*}{\textbf{\#}} & \multirow{2}{*}{\textbf{Vision}} & \multirow{2}{*}{\textbf{Instr.}} & \multicolumn{4}{c}{\textbf{Validation Unseen}}                 \\
                             &                                  &                                  & NE/m$\downarrow$ & SR/\%$\uparrow$ & OSR/\%$\uparrow$ & SDTW/\%$\uparrow$ \\ \hline
1                            & RGB+D                            & \checkmark                       & 122.7          & 4.5          & 13.9          & 1.0            \\
2                            & RGB                              & \checkmark                       & 145.5          & 3.2          & 7.9           & 1.0            \\
3                            & D                                & \checkmark                       & 205.5          & 3.0          & 18.1          & 0.8            \\
4                            & -                                & \checkmark                       & 177.0          & 1.3          & 12.1          & 0.3            \\
5                            & RGB+D                            & -                                & 145.4          & 2.1          & 11.1          & 0.5            \\ \hline
\end{tabular}
}
\vspace{-3mm}
\caption{Modality Ablations.}
\vspace{-8mm}
\label{tab:ablation}
\end{table}

It shows that both the vision and language inputs play the most important role (Row 1 \vs Row 4, Row 1 \vs Row 5). This is reasonable because without either of them the task actually is non-sense. Additionally, the large performance drop indicates the dataset has little visual or textual bias.
On the vision side, without depth information (Row 2) or RGB information (Row 3) leads to a success rate drop and without RGB drops more. This indicates both RGB and depth information matter to the final success and RGB information contributes more.

\section{Conclusion}
\label{sec:conclude}

In this work, we introduce a new task   and a large-scale dataset, AerialVLN, for the exploration of vision-and-language navigation in the sky. 
The linguistic analysis yields that AerialVLN dataset presents significant challenges to complex language understanding and its associated visual-textual alignment.
We also evaluate several widely adopted baselines, of which the performance drops significantly on our task and falls far behind human performance. 
This indicates that our task provides a broad study space for further research. 
\section{Acknowledgement}
This work was supported by National Key R\&D Program of China (No.2020AAA0106900), the National Natural Science Foundation of China (No.U19B2037), Shaanxi Provincial Key R\&D Program (No.2021KWZ-03), and Natural Science Basic Research Program of Shaanxi (No.2021JCW-03).

\clearpage

\newpage
{\small
\bibliographystyle{ieee_fullname}
\bibliography{egbib}
}

\end{document}